\documentclass{article}

\PassOptionsToPackage{numbers, compress}{natbib}
\usepackage[final]{nips_2017}

\usepackage[utf8]{inputenc} 
\usepackage[T1]{fontenc}    
\usepackage{hyperref}       
\usepackage{url}            
\usepackage{booktabs}       
\usepackage{amsfonts}       
\usepackage{nicefrac}       
\usepackage{microtype}      
\usepackage{graphicx}
\usepackage{amssymb}
\usepackage{amsmath}

\title{Video SemNet: Memory-Augmented Video Semantic Network}

\author{
  Prashanth Vijayaraghavan\\
  MIT Media Lab\\
  \texttt{pralav@mit.edu} \\
  \And
  Deb Roy \\
  MIT Media Lab \\
  \texttt{dkroy@media.mit.edu} \\
}

\begin{document}

\maketitle

\begin{abstract}
Stories are a very compelling medium to convey ideas, experiences, social and cultural values. Narrative is a specific manifestation of the story that turns it into knowledge for the audience. In this paper, we propose a  machine learning approach to capture the narrative elements in movies by bridging the gap between the low-level data representations and semantic aspects of the visual medium. 
We present a Memory-Augmented Video Semantic Network, called Video SemNet, to encode the semantic descriptors and learn an embedding for the video. The model employs two main components: (i) a neural semantic learner that learns latent embeddings of semantic descriptors and (ii) a memory module that retains and memorizes specific semantic patterns from the video. We evaluate the video representations obtained from variants of our model on two tasks: (a) genre prediction and (b) IMDB Rating prediction. We demonstrate that our model is able to predict genres and IMDB ratings with a weighted F-1 score of 0.72 and 0.63 respectively. The results are indicative of the representational power of our model and the ability of such representations to measure audience engagement. 
\end{abstract}

\section{Introduction}
The ubiquity of videos in this digital age and their ability to captivate the viewers have made them one of the strongest mediums of storytelling. However, not every video garners equal attention due to several factors like duration, content, target audience, etc. But narrative elements such as plot and structure play a crucial role in defining how well it will resonate with its audience. Research in video analysis \cite{lienhart1998comparison,li2004content,ekin2003automatic,naphade2002factor} revolving around narrative analysis, video semantic indexing and retrieval, summarization, etc. have focused on mapping low-level engineered features for understanding the video semantics. Most of these techniques use shot boundary detection, dominant color region detection or other shot detection methods to understand the aspects of the narrative content. Many times these low level features fail to capture the story or its narrative features. 

The recent advancements in the field of machine learning and deep learning have paved way for large-scale analysis of videos that include video classification \cite{karpathy2014large, zha2015exploiting}, text captioning \cite{pan2016hierarchical,yu2016video,shetty2015video} or summarization \cite{lu2013story, gong2014diverse, zhang2016video}, visual question answering \cite{antol2015vqa, fukui2016multimodal}, unsupervised video representation learning \cite{srivastava2015unsupervised, radford2015unsupervised, wang2015unsupervised}, etc. In this paper, we incorporate some of the ideas from previous studies and propose a new architecture that can learn latent semantic descriptors and remember some of the semantic patterns in a video using a memory module. The latent descriptors obtained from the video can provide a semantic summary of the video by a simple weighted summation. The memory module not only retains specific patterns in videos but also promotes transfer of information across various time steps. We leverage the ability of descriptors and memory module to generate a rich semantic representation for a given video. We train our models on a movie corpus (explained in section \ref{expt}) and experiment the representations produced by our model on genre and IMDB rating prediction tasks. These evaluations enable us to fathom how well our representations can (a) capture the video semantics, and (b) predict audience engagement. Our model can, therefore, bolster the efforts in improving the analysis around narrative aspects of a video. 

\begin{figure*}[t!]
\centering
  \includegraphics[width=\textwidth]{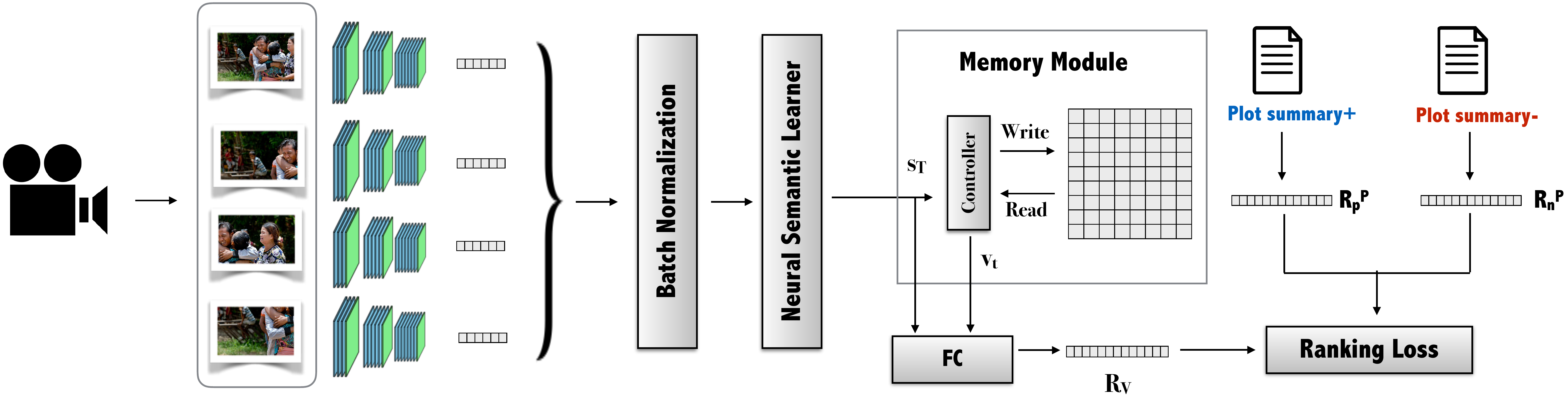}
  \caption{Illustration of Video SemNet}\label{semnet}
 \end{figure*}
\section{Video SemNet}
\paragraph{Overview:} Given a collection of movie videos, our goal is to leverage the structure and content of the videos to automatically learn a semantic representation that can encapsulate the narrative elements (like plot structure, setting, story, etc.) of the video and eventually help us predict audience engagement. The learning is driven by the positive and negative samples of plot summaries for each video. We introduce a memory-augmented video semantic network, henceforth referred to as Video SemNet, for this purpose. It comprises of two main components: (i) Neural Semantic Learner, and (ii) Memory Module. Figure \ref{semnet} illustrates the entire model. Before we go into the details of various components of our model, we briefly explain below the features and the loss function used to train our model. Video SemNet takes movie video as input and produces a video representation, $R_V$. Given both positive and negative samples of plot summaries related to the video, we train the model using a raking loss objective. 

\paragraph{Video Feature Extraction:} Motivated by Temporal Segment Network \cite{wang2016temporal}, our model enables learning over the entire movie by dividing it into $L$ equal segments and operating on a sequence of snippets (usually an RGB frame) sparsely sampled from the movie segments. We apply a pretrained convolutional architecture over those $L$ short snippets to extract rich latent features from the videos. A lot of literature \cite{simonyan2014very,yue2015beyond} in the past have demonstrated the capabilities of GoogleNet \cite{szegedy2015going}, Batch normalized-Inception \cite{krizhevsky2012imagenet}, ResNet-101 \cite{he2016deep}, etc. to extract features from still images. We use ResNet-101 on RGB frame snippets to get a 2048-dimensional feature vector $g_t$. Thus, the input feature vector can be represented as $x_t \in \mathbb{R}^{|g_t|}$ i.e $x_t \in \mathbb{R}^{2048}$. We complete this feature extraction step by a batch normalization component to deal with the problem of covariate shift. 
\paragraph{Modeling Plot summaries:} To compute the vector representation of positive and negative samples of plot summaries, we can utilize any of the encoding schemes \cite{mikolov2013distributed, pennington2014glove}. We adopt a two level positional encoding strategy. First, we encode every sentence in the plot summary individually using positional encoding. The positional encoding technique is applied on the resulting sentence encodings to produce an embedding for the entire plot summary. We implemented the positional encoding described in \cite{sukhbaatar2015end} instead of simple averaging of word vectors to prevent loss of sequence order in the plot summary. Formally, positional encoding for a sequence ($f_i$) is given by 
$
f_i =\sum_{j=1}^{L_s} {l_j \cdot w^j}
$
where $\cdot$ is element-wise multiplication and $l_j$ is a column vector with
structure $l_{jk} = (1 - j/L_s) - (k/d)(1 - 2j/L_s)$, where
k is the embedding index,  $d$ is the dimension of the
embedding, $w^j$ is an embedding for $j^{th}$ element (word or sentence) in the sequence and  $L_s$ is the length of the sequence. The embedding for the plot summary is denoted as $R^P$.

\paragraph{Loss Function:} 
We train our model to learn video representations using a ranking loss objective which measures their relevance to plot summaries. We want to maximize the similarity of video representation with its corresponding plot summary and minimize its similarity with negative examples. The objective function of the triplet ranking
loss is defined as
\begin{equation}
L=max(0,s(R_V, R^P_n) - s(R_V, R^P_p) + \alpha)
\end{equation}
where $\alpha$ is the margin parameter and $s(\cdot, \cdot)$ is the cosine similarity between video representation ($R_V$), positive ($R^P_p$) and negative ($R^P_n$) examples.

\subsection{Neural Semantic Learner}
Neural Semantic Learner builds a latent embedding for descriptors and outputs a semantic summary of the video. In order to the understand the narrative plot or structure from short snippets, it is important to track the variations between such sampled snippets. This can be accomplished by training these sequences of snippets through a recurrent neural network. Consider that we sample $L$ snippets $S \in \{S_1, S_2,..., S_L\}$ from their corresponding segments in the video. The features extracted for those sampled snippets are denoted as $X={x_1,x_2,....,x_L} $, where $X \in \mathbb{R}^{L \times 2048}$. The higher the value of $L$, the more computationally intensive the model turns out. Moreover, deeper LSTM do not always contribute to good performance as can be seen in \cite{yue2015beyond}. Hence, we introduce our input $X$  through temporal convolution layers with different filter sizes followed by a semantic descriptor learner. 
\paragraph{Temporal Convolutional Layers:} Let $L$ be the number of layers, $F_l$ is the number of convolutional filters at the layer $l$, and $K_l$ is the total time steps at layer $l$. We define a series of filters in each layer as $W=\{W^{(1)},W^{(2)},...,W^{(F_l)}\}$. With the output from previous layer $H^{(l-1)}$, we compute activations $H^{l}$ using
$
H^{l}=\Phi(W*H^{(l-1)}+b)
$
where $b$ is the the corresponding bias, $\Phi$ is the activation function. We perform a max pool operation that reduces the time step by half i.e. $K_l=K_{l-1}/2$.  Finally, the output from the temporal convolutional layers will be $H^{L}\in \mathbb{R}^{T_r \times d}$, $T$ is the resulting time steps and $d$ represents the dimensions of abstract video features. 

\paragraph{Semantic Descriptor Learner:} Semantic descriptor learner learns latent embeddings for each of the descriptors using a matrix $D \in \mathbb{R}^{N_D \times d}$ where $N_D$ is the number of descriptors, $d$ is the dimensions of video features.
The input $h_t$ to the learner refers a particular time step from $H^{L}$. Given $h_t$, we compute a weight vector $r_t$ over $N_D$ descriptors to measure its importance for that video. We want to have a function that can project $h_t$ to $N_D$ dimensions such that it models the temporal aspect of the video. Hence, we add a recurrence,
$
r_t=softmax(W_D [h_t;r_{t-1}])
$
where $W_D \in \mathbb{R}^{N_D\times (d+N_D)}$.
After computing $r_t$, we use it along with descriptor matrix $D$ to generate a semantic summarizer $s_t$ until time step $t$ as,
\begin{equation}
s_t=D^Tr_t
\label{summary}
\end{equation}
Finally, we use the semantic summary at time step $T$  ($s_T$) as input to our memory module.

\subsection{Memory Module}
Recently, memory-augmented neural network \cite{santoro2016one} has been applied in areas which require fast adaptation and memorizing events. An end-to-end differentiable
memory network is used to memorize facts that are relevant. In our scenario of video semantic analysis, we take inspiration from previous work associated with memory networks \cite{graves2014neural, kaiser2017learning} as it helps reason with the limited data and improve our ability to represent critical, unseen instances of the video. Our memory module comprises of (i) an external memory matrix $M  \in \mathbb{R}^{m \times d}$, were $m$ is the size of the memory and $d$ is the dimensionality of each location in the memory (ii) a neural controller which uses a read head $w^r_t$ and age vector $A_t$ (${w^r_t, A_t} \in \mathbb{R}^m$) to perform memory operations at a particular time step $t$. The various memory operations are summarized below:
\paragraph{Memory Read:} The read head $w^r_t$ is computed from semantic summary ($s_T$), memory state ($M_t$) and controller context $C_t$ via,
$w^r_t(i)=softmax(\theta_t(i))$ where $\theta_t(i)=tanh(W_s^\alpha s_T+W_c^\alpha C_{t}+W_m^\alpha M_t(i))$ and $W_s^\alpha, W_c^\alpha, W_m^\alpha, W_h^\alpha$ are trainable parameters, $C_{t}$ is the  controller context vector that summarizes the controller operations until time step $t$. We get the memory read vector using $v_t=M_t(j)$ where $j=argmax_i(w^r_t(i))$. After the read step, we perform two operations: (1) a new controller context vector ($C_{t+1}$) is calculated using a non-linear combination of current context vector ($C_t$), current read vector $v_t$ and our semantic summary $s_T$ via $C_{t+1}=tanh(W_c^\beta C_t+ W_v^\beta v_t+ W_s^\beta s_T)$, where $W_s^\beta, W_c^\beta, W_h^\beta$ are learnable parameters, and (2) update age vector $A_t$ by incrementing the age of all indices by 1, i.e., $A_{t+1}=A_t+1$.
\paragraph{Memory Update:} The memory update is a linear projection of the updated context vector to a particular memory location $p$, given by, $M_{t+1}(p)=W_m^\gamma C_{t+1}$, where $W_m^\gamma$ is a learnable projection matrix. We find memory location $p$ by selecting the index with maximum age in age vector $A_t$ with some randomness. Formally, we choose the memory location $p$ to update via $p=argmax_i(A_{t+1}(i))+r$ where $r \ll m$ is a random number so as to avoid race conditions in asynchronous training. Finally, we refresh the age vector by setting $A_{t+1}(p)$ to 0. Our update mechanism, that conditions controller's context vector $C_t$ with read vector $v_t$, promotes the flow of extracted semantic information across various time steps. During an update, whenever there is a need to remember the contents of memory location that is being
overwritten, the model training will enforce the controller to copy its read vector to the
corresponding memory location. 

The final representation that encodes the semantic representation for the video is computed using a combination of both summary vector $s_T$ and memory read vector $v_t$, formally given by,
\begin{equation}
R_V=tanh(W_{sv} [s_T;v_t])
\end{equation}
where $W_{sv} \in \mathbb{R}^{d \times 2d}$ represents the trainable projection matrix.

\section{Experiments}
\label{expt}
Since our model learns the representations from plot summaries, the choice of our dataset is movies. The movies dataset comprises of 1436 Hollywood films available on DVD. We performed data augmentation as mentioned in \cite{wang2016temporal} for our training. We train different variants of our model on this dataset: (i) Simple Standard Model (SSM): model without neural semantic learner and memory module ($R_v$ is a projection over mean of extracted video features), (ii) Semantic Learner Model (SLM): model includes neural semantic learner but no memory module, (iii) Video SemNet: model includes both semantic learner and memory module. In order to evaluate the video representations, we use them for two tasks (a) movie Genre Prediction, (b) IMDB Rating Prediction. Each movie in IMDB website \footnote{http://www.imdb.com/} is associated with genres and ratings. For genre prediction task, we shortlisted $\sim$100 movies for each of the five most frequent genres: action, comedy, drama, horror and romance. We used 80\% of data for training our model using the learned representations and tested it on the remaining 20\%. For IMDB rating prediction, we round off the ratings associated with movies and classify them across 10 different classes (Ratings 1-10) with baseline accuracy of 30\%. Table \ref{eval} shows the results of our evaluation. We find that our Video SemNet model outperforms the other models on both the tasks. 

\begin{table}[t]
 
  \centering
  \begin{tabular}{ccc}
    \toprule
    \cmidrule{1-2}
    Models     & IMDB Rating Prediction     & Genre Prediction \\
    \midrule
    SSM & 0.48 & 0.54    \\
    SLM     & 0.58 & 0.62      \\
    Video SemNet     & \textbf{0.63}    & \textbf{0.72}  \\
    \bottomrule
  \end{tabular}
   \caption{Evaluation of different variants of our model on two tasks: Genre Prediction and IMDB Rating Prediction. We show that Video SemNet outperforms all the other models based on weighted F1-score.}
  \label{eval}
\end{table}
\section{Conclusion}
We have proposed a new model to compute a semantic embedding for videos that can better represent the narrative aspects of a video. 
The representations obtained from our model are able to predict movie genres and IMDB ratings successfully, proving such representations can be generalized across different tasks. Though the current model showcases the power of our representations, it would be interesting to experiment with incremental improvements to the model like adding optical flow features and combining  emotional arc analysis \cite{chu2017audio} with our semantic analysis. Beyond audience engagement analysis, we can also use such representations to generate summaries, predict propagation patterns on social media platforms, perform video retrieval and a lot more.

\subsubsection*{Acknowledgments}
The authors would like to thank Eric Chu for his help with data collection for this project.


\end{document}